\documentclass{article}
\usepackage{spconf,amsmath,graphicx}
\usepackage{amssymb}
\usepackage{booktabs}

\title{Attention-Based LSTM for Psychological Stress Detection from Spoken Language Using Distant Supervision}
\name{Genta Indra Winata, Onno Pepijn Kampman, Pascale Fung\thanks{This work is partially funded by CERG16214415 of the Hong Kong Research Grants Council and ITS170 of the Innovation and Technology Commission.}}
\address{
  Human Language Technology Center\\
  Department of Electronic and Computer Engineering\\
  Hong Kong University of Science and Technology, Clear Water Bay, Hong Kong\\
  \{giwinata, opkampman\}@connect.ust.hk, pascale@ece.ust.hk
}

\begin{document}
%
\maketitle
\begin{abstract}
We propose a Long Short-Term Memory (LSTM) with attention mechanism to classify psychological stress from self-conducted interview transcriptions. We apply distant supervision by automatically labeling tweets based on their hashtag content, which complements and expands the size of our corpus. This additional data is used to initialize the model parameters, and which it is fine-tuned using the interview data. This improves the model's robustness, especially by expanding the vocabulary size. The bidirectional LSTM model with attention is found to be the best model in terms of accuracy (74.1\%) and f-score (74.3\%). Furthermore, we show that distant supervision fine-tuning enhances the model's performance by 1.6\% accuracy and 2.1\% f-score. The attention mechanism helps the model to select informative words.
\end{abstract}
\section{Introduction}
\label{sec:intro}
Psychological stress has a serious effect on mental health and is often a precursor for more severe conditions. Although stress is a natural stimulant, persistent increased levels yield adverse effects, such as heart attacks \cite{yudkin00}, hypertension \cite{matthews04}, and addiction \cite{slopen13}. Prolonged stress is also linked to mental health issues like anxiety \cite{faravelli89} and depression \cite{breslau95}. Its prevalence has been increasing in the past decade \cite{APA17} and affects the way people speak and their choice of spoken language. Emotional support is known to alleviate stress, yet less than 50\% of the stressed population receives enough support from friends, family and professionals \cite{APA17}. Linguistic studies have shown that language choice contains pointers to levels of stress and mental health \cite{o2017linguistic}. The potential of text data from social media and Twitter for predicting major depression occurrence has also been demonstrated~\cite{de2013predicting, cavazos2016content}.



Research on sentence-level stress detection has been mostly focused on written text collected from social media such as micro-blogs \cite{lin2014psychological, lin2016does}. The authors of these two works used a framework that combines linguistic, visual and social attributes in classifying stress categories. Lin et al. explored tweets to find stressors and stressful events, by building a stressors and stress subject dictionary. They collected thousands of written Weibo tweets and manually categorized them into 10 groups \cite{lin2016does}. On word-level stress detection, a simple bidirectional RNN can achieve good results on Russian speech transcriptions \cite{ponomareva2017automated}.

In this work, we propose to build attention-based Long Short-Term Memory (LSTM) models fed with word embeddings for detecting psychological stress on sentence-level from interview transcriptions. It takes a long dependency across words in an utterance. Then, the attention mechanism weighs the importance of every word and chooses what to retrieve from the memory. It outputs the weighted combination of all words to the network for predicting the stress level. We apply distant supervision by adding unlabeled tweets from Twitter to our training set. This technique refers to extracting noisy signals from text as label \cite{marchetti2012learning}. In our case, we manually pick hashtags that indicate either a stressed or unstressed state of mind of the author, and use them to scrape stressed (positive labels) and unstressed (negative labels) tweets. We need to include more data during training because our interview corpus is relatively small and covers a limited number of topics, mostly related to academia. The major contribution of this paper is to show that unlabeled data collected from Twitter can improve the classification performance on our interview transcriptions corpus, and that applying an attention mechanism helps the model to effectively choose important words. 


\section{Methodology}
\label{sec:methodology}

Our objective is to determine whether someone is stressed or not, given an utterance as input. We explore several different models. For the LSTM and bidirectional LSTM (BLSTM) models, we use a trainable embedding layer whose vectors should eventually form stressed and unstressed term clusters. LSTMs can capture temporal dynamics of words in a sentence.

\subsection{Long Short-Term Memory (LSTM)}
First, we build a unidirectional LSTM \cite{hochreiter1997long} taking word embedding as input. We denote $V$ as the number of unique words in our corpus and $k$ as the dimension of the word embedding vectors. Each word is a one-hot vector $\boldsymbol{x}$ $\in$ $\mathbb{R}^{|V|}$ and performs a multiplication with the embedding layer $A$ $\in$ $\mathbb{R}^{|V| \cdot k}$, where $k = 100$. The resulting vector is $\boldsymbol{b}$ $\in$ $\mathbb{R}^{k}$


\begin{equation}
	\boldsymbol{b} = A^{T}\boldsymbol{x}
\end{equation}

\begin{figure}[t]
  \centering
  \includegraphics[width=0.74\linewidth]{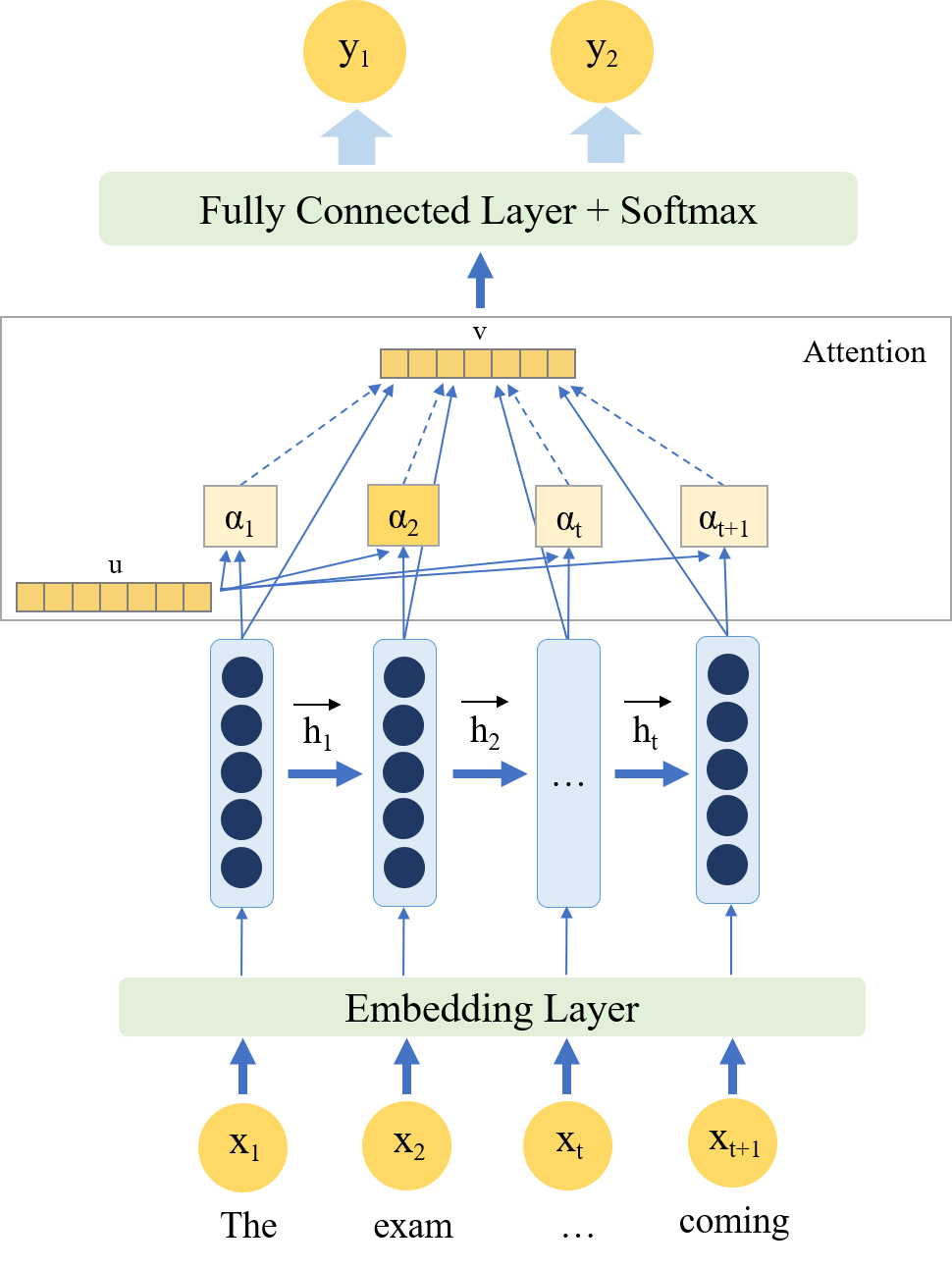}
  \caption{Attention-based LSTM architecture}
  \label{fig:lstm_architecture}
\end{figure}

The model architecture is shown in Figure~\ref{fig:lstm_architecture}. The LSTM consists of one recurrent layer that propagates the embedding vector $b_t$ for the word at time $t$ (i.e. a column of $\boldsymbol{b}$) through the LSTM network to find hidden state $h_t$

\begin{equation}
	\overrightarrow{h_t} = LSTM(b_t), t \in [1, T]
\end{equation}

\begin{figure}[th]
  \centering
  \includegraphics[width=0.74\linewidth]{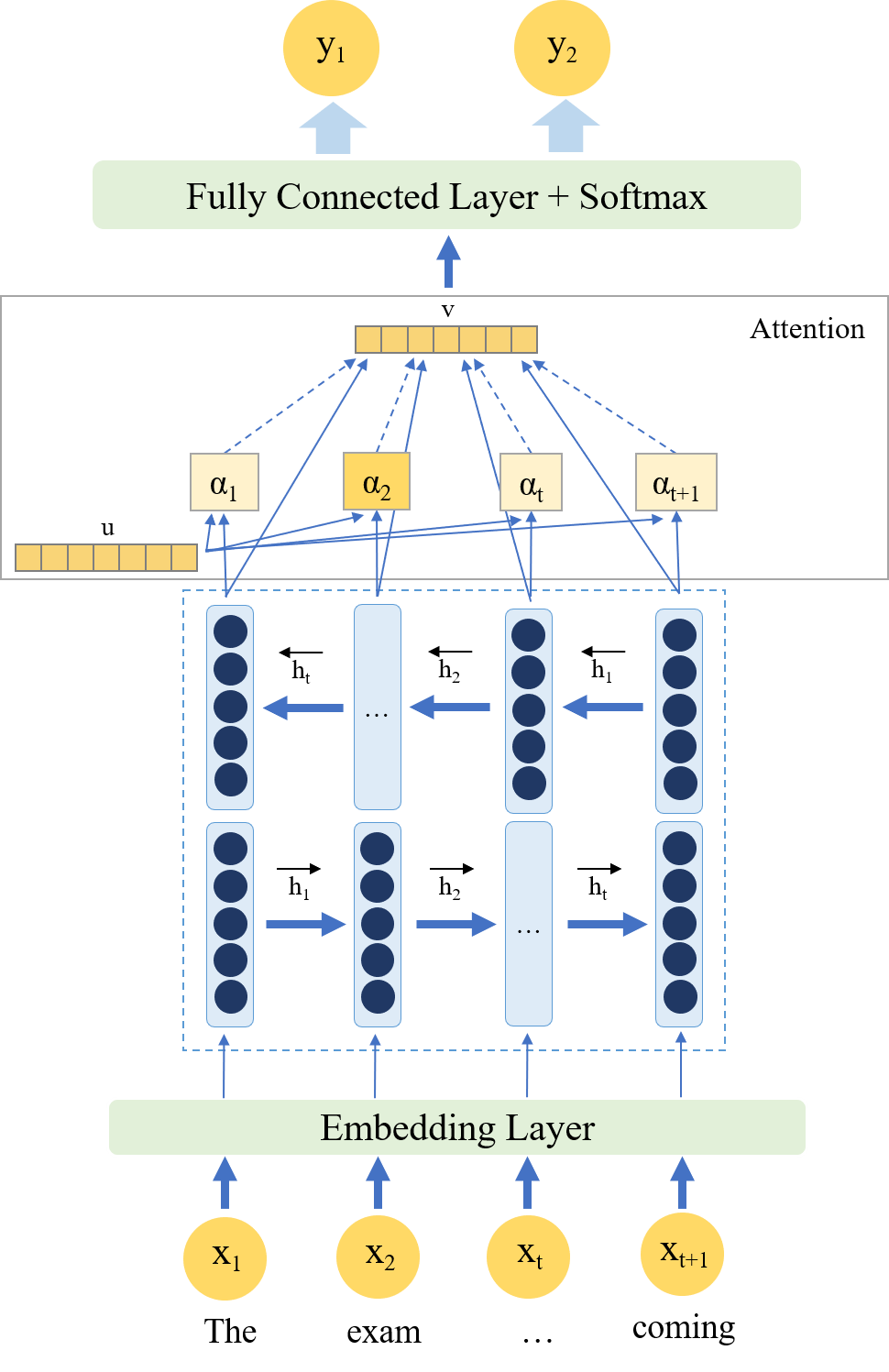}
  \caption{Attention-based BLSTM architecture}
  \label{fig:bi_lstm_architecture}
\end{figure}

All hidden states are fed into a subsequent attention layer~\cite{yang2016hierarchical}. We added this layer because not all words contribute equally to the stress classifier. The word importance vector $u_t$ is calculated with Equation~\ref{eq:context_vector}. The normalized word weight $\alpha_t$ is obtained through a softmax function (Equation~\ref{eq:word_weight}). The aggregate of all the information in the sentence $\textbf{v}$ is the weighted sum of each $h_t$ with $\alpha_t$ as corresponding weights.

\begin{equation}
\label{eq:context_vector}
	u_t = tanh(W \overrightarrow{h_{t}} + b)
\end{equation}

\begin{equation}
\label{eq:word_weight}
	\alpha_{t} = \frac{exp(u_{t}^Tu)}{\sum_{t}exp(u_{t}^T u)}
\end{equation}

\begin{equation}
	\boldsymbol{v} = \sum_t \alpha_{t}\overrightarrow{h_{t}}
\end{equation}

This vector $\textbf{v}$ is then fed to a fully connected layer with softmax activation to perform the final classification. The prediction is a vector $\boldsymbol{y}$ $\in$ $\mathbb{R}^2$ with the probabilities of being unstressed and stressed. We choose the highest probability by using $argmax$ as the model's prediction.

\subsection{Bidirectional Long Short-Term Memory (BLSTM)}
We train a BLSTM model in identical fashion to process the word sequence in both forward ($\overrightarrow{h_{t}}$) and backward direction ($\overleftarrow{h_{t}}$). This recurrent neural network uses two LSTMs, one for each direction. The architecture is shown in Figure~\ref{fig:bi_lstm_architecture}.





\subsection{Support Vector Machine (SVM)}
As a baseline, we build an SVM \cite{cortes1995support} with a Radial Basis Function (RBF) kernel. We extract {\tt word2vec} \cite{mikolov2013distributed} vector embeddings for each word in a given sentence. The embeddings have dimensionality of $k=300$ and were pre-trained on Google News data (around 100 billion words with around 3 million unique words). Since the utterances have a variable number of words, we compute the average sentence vector for each embedding feature and train on the remaining feature space. We average the sum of all vectors to get a new word embedding vector, such that

\begin{equation}
	b_j = \frac{\sum_{i=<N>}{a_{i, j}}}{N}
\end{equation}

where $a_{i, j}$ is the word embedding vector of word $i$ in sentence $j$, and $b_{j}$ is the sentence vector. Thus, for the SVM, the input is represented as an input matrix consisting of $N$ utterance vectors.

\section{Experiments}
\label{sec:experiment}

\begin{table} [t]
	\centering
    \small
	\caption{Twitter hashtags.}
    \label{tab:twitter-hashtags}
  \begin{tabular}{|p{8cm}|}
  \hline
  \textbf{\# stressed}
   \\ \hline
amstressed, busylife, collegestress, distress, distressed, familystress, feelingbusy, feelingfrustrated, feelingoverwhelmed, feelingstress, feelstress, feelstressed, frustrated, frustrating, frustration, iamstressed, ifrustrated, imstressed, overwhelm, overwhelmed, overwhelming, panic, sostress, sostressed, sostressful, stress, stressed, stressedlife, stressedout, stresses, stressful, stressfulllife, stressingout, stresslife, stressor, stressors, stresss, stressss, stresssss, verystressed, workstress \\ \hline
\textbf{\# unstressed} \\ \hline
blessed, comfort, feelingrelax, feelingrelaxed, grateful, iamblessed, iamgrateful, iamrelaxed, imblessed, imgrateful, nostress, peaceful, relax, relaxed, relaxing\\ \hline
  \end{tabular}
\end{table}

\begin{table}[t]
\centering
\small
\caption{Interview Corpus statistics.}\label{tab:interview-corpus}
	\begin{tabular}{|c|c|c|c|}
	\hline
    \textbf{utterances} & \textbf{tokens} & \textbf{speakers} & 	\textbf{vocab size}\\ \hline
    2,272 &36,538 &38 &3,127\\ \hline
	\end{tabular}

\bigskip

\caption{Twitter Corpus statistics.}\label{tab:twitter-corpus}
	\begin{tabular}{|c|c|c|c|c|}
	\hline
    \textbf{tweets} &\textbf{tokens} &\textbf{stressed} &\textbf{unstressed} &\textbf{vocab size} \\ \hline
    367,312 &5,439,427 &59,768 &307,544 &135,463 \\ \hline
	\end{tabular}
\end{table}



\subsection{Corpora}
For our experiments, we used two different corpora: an interview corpus, the Natural Stress Emotion corpus~\cite{zuo2012multilingual}, and the Stress Twitter Corpus. 
The former corpus contains 25 students (13 females) answering the same set of 12 interview questions. The questions were designed to be progressively stress provoking. Additionally, we expanded the dataset by conducting 13 more interviews (3 with females) with identical setup. All answers were binary labeled for stress by three judges, from which we took the majority vote. It has four hours of recordings in total with 36,538 word tokens (see Figure~\ref{tab:interview-corpus}). For the text-based models described here, we only used the English transcriptions. Because this corpus is small, we collected more data from Twitter and selected tweets with a set of filtering heuristics based on \cite{wang2012harnessing}. We only kept tweets with the hashtag at the end, and having less than four hashtags in total. We filtered out tweets containing URLs and images, and applied distant supervision ~\cite{marchetti2012learning} to label  the unlabeled tweets. That is, we manually chose hashtags that indicate either stressed or unstressed state of mind of the author (See Figure ~\ref{tab:twitter-hashtags}), and used these to automatically label our scraped tweets. Not all text is created equally and it's important here to be aware of the differences between spoken language and written language on social media.

\subsection{Setup}
\label{ssec:setup}


For the LSTM and BLSTM experiments, the recurrent layer consists of 64 units. In order to regularize the model, a dropout layer \cite{srivastava2014dropout} with probability of 0.2 is inserted between the recurrent and attention layers. We use batch gradient descent using Adam \cite{kingmaadam} as optimizer, with batches of 128 samples. We also run both the LSTM and BLSTM without attention mechanism for comparison.

We take 160 random samples from each class from the interview corpus as our test set, as we want to evaluate on spoken language. The remainder is used as training set.  All sentences are padded to 35 words. In order to balance the distribution of our training set, we oversample the minority class (stressed) within the training set. We validate the model to find the best setting. For two iterations, the model is trained only with twitter data and afterwards, the model subsequently fine-tuned with interview data. Twitter data is inherently different from spoken transcripts, and both are noisy (absence of correct grammar) in their own way. Since the Twitter corpus is imbalanced, we random sample and take 49,000 tweets from each class every iteration.



\begin{table} [t]
	\centering
    \small
	\caption{Model performance.}
    \label{tab:results}
    \begin{tabular}{|p{2.8cm}|c|c|c|c|}
    \hline
    \textbf{method} & \textbf{accu.} & \textbf{prec.} & \textbf{recall} & \textbf{f-score} \\ \hline
    SVM &68.7 &72.0 &61.2 &66.2 \\ \hline 
    LSTM &70.0 &70.3 &68.1 &69.2 \\ \hline
    LSTM w/ attention &\textbf{73.8} &\textbf{74.7} &\textbf{71.9} &\textbf{73.2} \\ \hline
    BLSTM &72.2 &74.5 &67.5 &70.8 \\ \hline
    BLSTM w/ attention &72.5 &73.1 &71.2 &72.2 \\ \hline
	\end{tabular}
\end{table}

\begin{table}[t]
  \centering
  \small
  \caption{Fine-tuning performance.}
  \label{tab:fine-tuned-results}
  \begin{tabular}{|p{2.8cm}|c|c|c|c|}
  \hline
  \textbf{method}  & \textbf{accu.} & \textbf{prec.} & \textbf{recall} & \textbf{f-score} \\ \hline
  LSTM               & 73.4             & 73.6             & 73.1              & 73.4               \\ \hline
  LSTM w/ attention  & 73.8             & 74.4             & 72.5              & 73.4               \\ \hline
  BLSTM              & 73.8             & \textbf{74.7}             & 71.9              & 73.2               \\ \hline
  BLSTM w/ attention & \textbf{74.1}             & 73.6             & \textbf{75.0}              & \textbf{74.3}               \\ \hline
  \end{tabular}
\end{table}

\subsection{Results}
\label{ssec:result}
Relevant evaluation results on the test set are shown in Tables~\ref{tab:results} and~\ref{tab:fine-tuned-results}. The performance of the BLSTM with attention outperforms the other classifiers in terms of accuracy and f-score. The fine-tuning process helps the model to classify sentences related to stress better, but significantly increases the recall. Models with attention mechanism are slightly better. 

\subsection{Discussion}
\begin{figure}[t]
  \centering
  \includegraphics[width=\linewidth]{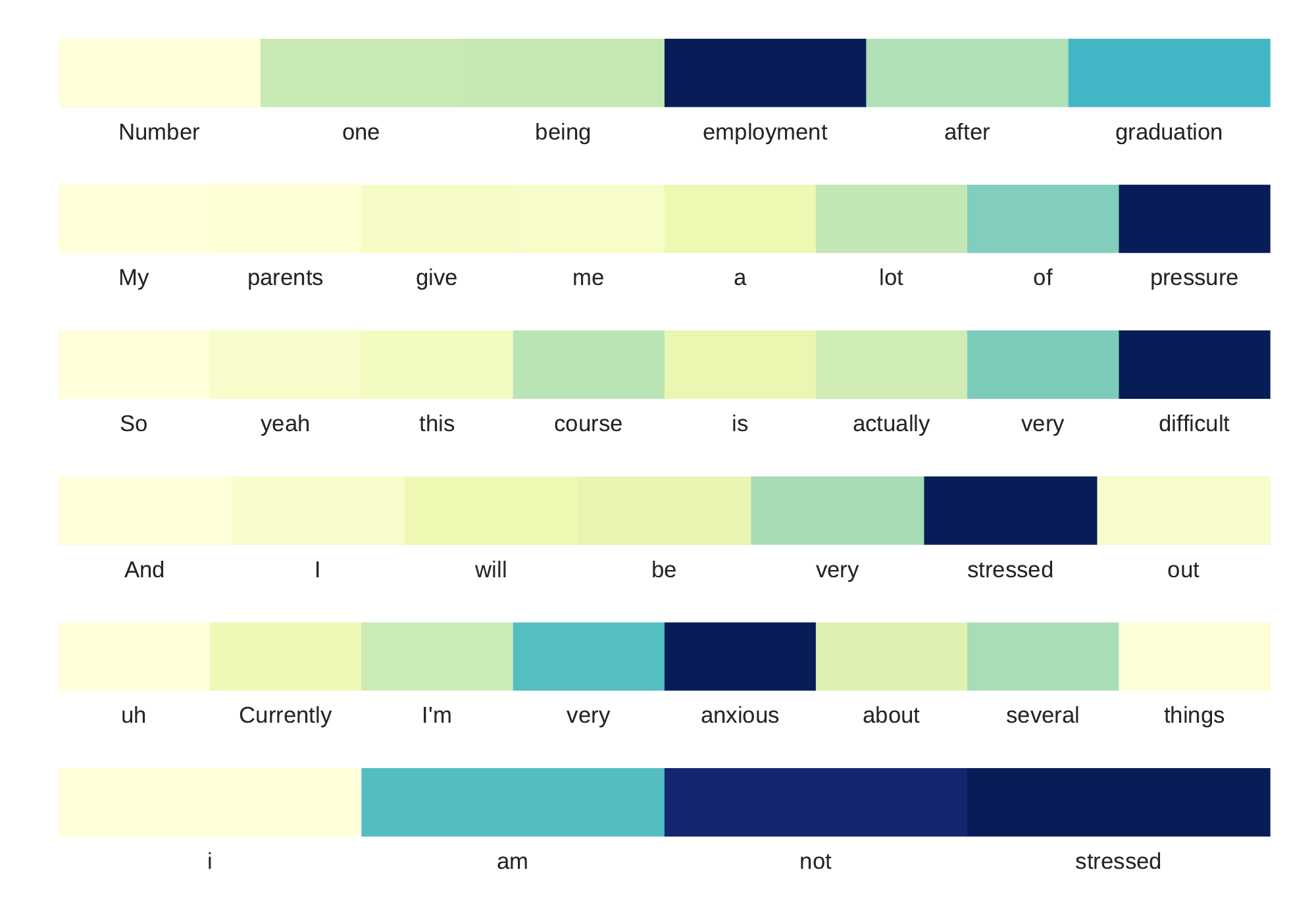}
  \caption{Heatmap of attention layer weights for stressed utterances.}
  \label{fig:stressed-utterances}
\end{figure}

\begin{figure}[t]
  \centering
  \includegraphics[width=\linewidth]{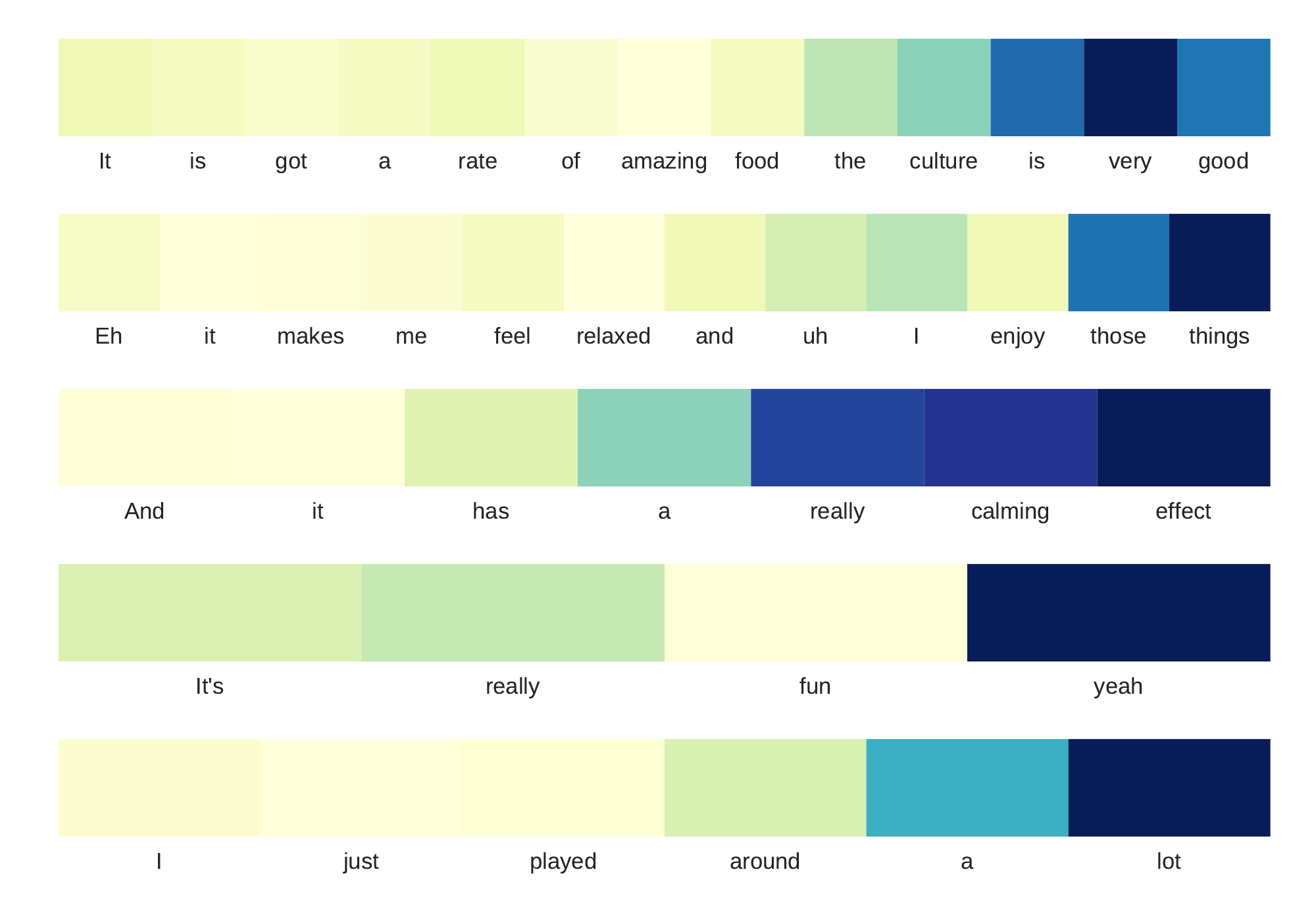}
  \caption{Heatmap of attention layer weights for unstressed utterances.}
  \label{fig:unstressed-utterances}
\end{figure}

To visualize the attention mechanism, we extract the attention weights from the best trained model and evaluate several stressed and unstressed utterances (see Figures~\ref{fig:stressed-utterances} and ~\ref{fig:unstressed-utterances}). The figures show the contribution of each word in the classification task. Darker colors represent stronger word contributions to the classification task. Interestingly, the added attention layer captures key terms related to stress. For instance, in the first stressed example, words such as ``employment" and ``graduation" weigh heavier than others. These words have stronger relation with stress. Furthermore, we can see from other stressed samples that words such as ``employment", ``pressure", ``difficult", ``stressed", and ``anxious" have similar weights. Conversely, words such as ``my", ``Number", ``I", and ``And" are least considered for the classification because they are not related to stress.

Distant supervision seems to slightly improve the performance, especially the recall (i.e. false negatives are turned to true positive). This is likely due to the interview corpus covering a limited domain (mainly academic issues), and all interviewees answering the same questions. The domain of the Twitter corpus is more general because it includes other domains and is approximately 40 times larger than the interview corpus, thus adding it makes the model more robust. However, the model did not learn more complicated grammatical structures. For example, from Figure~\ref{fig:stressed-utterances} we can see that "I am not stressed" is classified as stressed. The observation that the model does not learn the semantic meaning of negation, could be caused by a lack of data. Also, we believe tweets are not a great source for models to learn proper grammatical structures.

Our models learn statistical characteristics of language choice under stressed psyche. That is, they learn which word combinations and sequences are expressed more often when someone is stressed. It is showed more explicitly in Figure~\ref{fig:stressed-utterances}. Obviously, a person can talk about stressful topics, yet still remain calm, and vice versa. Although there is an obvious signal, not all stress information is encoded in language choice. This inherently limits our model. The interview corpus only contains transcribed spoken language. A more complete stress detection framework would also include context and prosodic features of spoken utterances. 




\section{Conclusion}
We have presented methods for classifying interviewee stress level from interview transcriptions. The best performance was found for our bidirectional LSTM model, which outperformed the other models in terms of accuracy and f-score. The two-phase training method with the out-of-domain stress tweets dataset improves the learning performance. Future work includes multi-modal learning using linguistic and acoustic features. We are also interested in gathering more grammatically correct sentences for transfer learning purposes, so the model may learn how to deal with negation (among others). Furthermore, we will incorporate the model described here into our virtual therapist platform~\cite{winata2017nora}, where it is fed with Automatic Speech Recognition output. This makes the system aware of user stress, to which it responds with appropriate stress management advice and exercises. 


\bibliographystyle{IEEEbib}
\bibliography{strings,refs}

\end{document}